\documentclass[runningheads]{llncs}
\usepackage{graphicx}
\usepackage{subfigure}
\usepackage{multirow}
\usepackage{comment}
\usepackage{color}
\usepackage{marvosym}
\usepackage{amsmath}
\begin{document}
\title{EfficientSRFace: An Efficient Network with Super-Resolution Enhancement for Accurate Face Detection\thanks{Supported by the Natural Science Foundation of China (NSFC) under grants 62173186 and 62076134.}}
%
%

\titlerunning{EfficientSRFace}

\author{Guangtao Wang\inst{1} \and
Jun Li\inst{1}\textsuperscript{\Letter} \and
Jie Xie\inst{1} \and
Jianhua Xu\inst{1} \and
Bo Yang\inst{2}
}

\authorrunning{G. Wang et al.}
%

\institute{School of Computer and Electronic Information, Nanjing Normal University, 210023, China \and
School of Artificial Intelligence, Nanjing University of Information Science and Technology, 210044, China \\
\email{\{202243023,lijuncst,73049,xujianhua\}@njnu.edu.cn, 003402@nuist.edu.cn}}

%
\maketitle              
\begin{abstract}
In face detection, low-resolution faces, such as numerous small faces of a human group in a crowded scene, are common in dense face prediction tasks. They usually contain limited visual clues and make small faces less distinguishable from the other small objects, which poses great challenge to accurate face detection. Although deep convolutional neural network has significantly promoted the research on face detection recently, current deep face detectors rarely take into account low-resolution faces and are still vulnerable to the real-world scenarios where massive amount of low-resolution faces exist. Consequently, they usually achieve degraded performance for low-resolution face detection. In order to alleviate this problem, we develop an efficient detector termed EfficientSRFace by introducing a feature-level super-resolution reconstruction network for enhancing the feature representation capability of the model. This module plays an auxiliary role in the training process, and can be removed during the inference without increasing the inference time. Extensive experiments on public benchmarking datasets, such as FDDB and WIDER Face, show that the embedded image super-resolution module can significantly improve the detection accuracy at the cost of a small amount of additional parameters and computational overhead, while helping our model achieve competitive performance compared with the state-of-the-arts methods.

\end{abstract}
\section{Introduction}\label{sec1}

With the development of deep convolutional neural networks (CNN), dramatic progress has been made recently in face detection which is one of the most fundamental tasks in computer vision~\cite{crface,pyramidbox,groupface}. With superior representation capability, deep models have achieved unrivaled performance compared with traditional models. In pursuit of high performance, particularly, numerous heavyweight face detectors~\cite{mogface,alnnoface,srnface} are designed with excessive parameters and complex architecture. e.g., the advanced DSFD detector~\cite{dsfd} has 100M+ parameters, costing 300G+ MACs. Although various lightweight designs are used for producing simplified and streamlined networks~\cite{yolo5face,extd,lffd}, the models trading accuracy for efficiency suffer from degraded performance. Recently, more efforts are devoted to designing efficient network, and the EfficientFace detector~\cite{efficientface} has been proposed recently for addressing the compromise between efficiency and accuracy.

\begin{figure}
\centering
\includegraphics[scale=0.5]{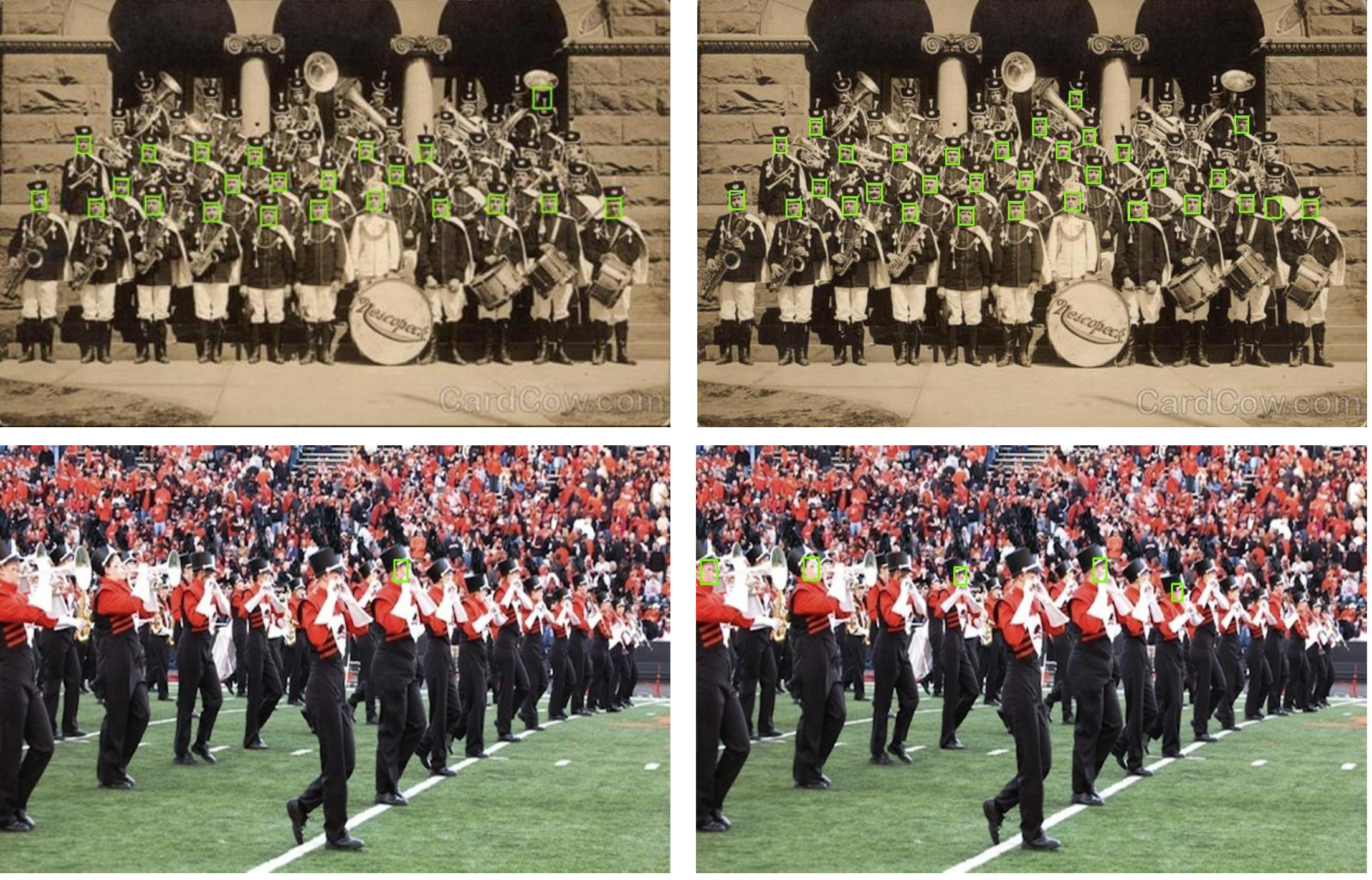}
\caption{Visualization of detection results achieved by EfficientFace (left) and our EfficientSRFace (right) in two different images. The results demonstrate the advantage of our network against the EfficientFace in the case of dense low-resolution face detection.}
\label{fig11}
\end{figure}


However, in real-world scenarios, low-resolution faces account for a large proportion in dense face detection tasks. For example, massive amount of small faces of a human group exist in a crowded scene for a low-quality image. They usually contain limited visual clues, making it difficult to accurately distinguish them from the other small objects and posing great challenge to accurate detection. Although current deep detectors have achieved enormous success, they are still prone to low low-resolution face detection accuracy. As shown in Fig.~\ref{fig11}, when handling the images including a large amount of densely distributed faces of a large human group with considerable variances, the EfficientFace reveals deteriorating performance without correctly identifying the low-resolution faces (Some small-scale blurred faces in the images are missing in the detection results).  To alleviate this problem, we embed a feature-level image super-resolution reconstruction network into EfficientFace, and design a new detection framework termed EfficientSRFace for improving the accuracy of low-resolution face detection. As a simple residual attention network, the newly added reconstruction module can enhance the feature representation capability of our detector at the cost of a small amount of additional parameters and limited computational overhead growth. Notably, the module is only introduced into the training process and is discarded for inference, and thus inference efficiency is not affected. 


To summarize, our contributions in this study are twofold as follows:

\begin{itemize}
   \item In this paper, we develop a new efficient face detection architecture termed EfficientSRFace. Based on the well-established EfficientFace, a feature-level image super-resolution network is introduced, such that the feature representation capability of characterizing low-resolution faces is enhanced.
   \item Extensive experiments on public benchmarking datasets show that the super-resolution module can significantly improve the detection accuracy at the cost of a small amount of additional parameters and computational consumption, while helping our model achieve competitive performance compared with the state-of-the-arts.
\end{itemize}


\section{Related Work}\label{sec2}

\subsection{Face Detection}\label{sec21}
With the rapid development of deep networks for general-purpose object detection~\cite{ssd,yolo,centernet,efficientdet}, significant progress has been made in face detection. Recently, various heavyweight face detectors have been designed to realize accurate face detection~\cite{fduifr,isrnet,rahpfd,tinaface}. In order to speed up computation and reduce network parameters, Najibi et al.~\cite{ssh} proposed SSH detector by utilizing feature pyramid instead of image pyramid and removing the fully connected layer of the classification network. Tang et al.~\cite{pyramidbox} proposed a new context assisted single shot face detector termed Pyramidbox considering context information. Liu et al.~\cite{hambox} designed HAMBox model which incorporates an online high-quality anchor mining strategy that can compensate mismatched faces with high-quality anchors. In addition, ASFD~\cite{asfd} combines neural structure search techniques with a newly designed loss function. Although the above models have superior performance, they have excessive architectural parameters and incur considerable costs during the training process. Lightweight model design has become the promising line of research in face detection. One representative lightweight model is EXTD~\cite{extd}, which is an iterative network sharing model for multi-stage face detection and significantly reduces the number of model parameters. Despite the success of both heavyweight and lightweight detectors, they still suffer insufficient descriptive ability of capturing low-resolution face, and thus reveal inferior performance when handling low-resolution face detection in real-world scenarios.

\subsection{CNNs for Image Super-Resolution}\label{sec22}
Benefiting from the promise of CNN, major breakthroughs have also been made in the field of super-resolution (SR) reconstruction. In 2015, Dong et al.~\cite{isdcn} proposed a deep learning framework for single image super-resolution named SRCNN. For the first time, convolutional networks were introduced into SR tasks. Later, they improved SRCNN and introduced a compact hourglass CNN structure to realize faster and better SR model~\cite{fsrcnn}. In order to improve the performance of image reconstruction, Kim et al.~\cite{aisdcn} proposed a deeper network model named VDSR. By cascading small filters in the deep network structure multiple times, the context information can be effectively explored. Wang~\cite{esrgan} developed ESRGAN and reduced computational complexity by removing the Batch Normalization (BN) layer and adding a residual structure. Zhang et al.~\cite{rcan} proposed a very deep residual channel attention network, which integrates the attention mechanism into the residual block and forms the residual channel attention module to obtain high-performance reconstructed images. ClassSR~\cite{classsr} proposed SR pipeline that combined classification and super-resolution on the sub-image level and tackled acceleration via data characteristics. Cong et al.~\cite{hrihcdt} unified pixel-to-pixel transformation and color-to-color transformation coherently in an end-to-end network named CDTNet. Extensive experiments demonstrate that CDTNet achieved a desirable balance between efficiency and effectiveness. In this paper, in order to alleviate the drawback of existing face detectors in low-resolution face detection, an image super-resolution network is embedded into our EfficientFace network to enhance the feature expression ability of the model. To our knowledge, this is the first attempt to incorporate the SR network into efficient face detector to address low-resolution face detection.

\section{EfficientSRFace}\label{sec3}

In this section, we will briefly introduce our proposed EfficientSRFace framework followed by a detailed description of the embedded feature-level super-resolution module. In addition, the loss function of our network will also be discussed.

\subsection{Network Architecture}\label{sec31}

\begin{figure}
\centering
\includegraphics[scale=0.4]{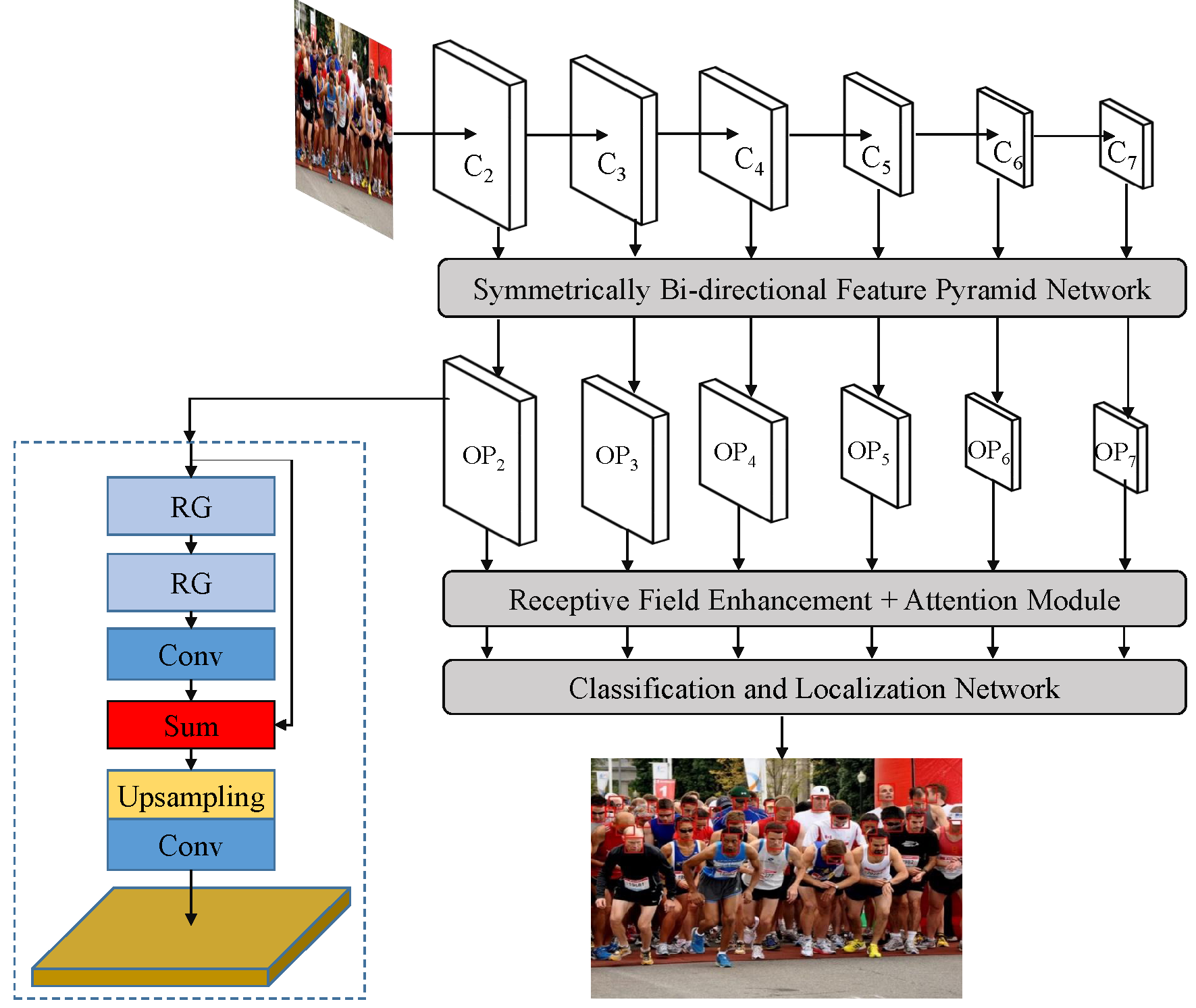}
\caption{The network structure of our proposed EfficientSRFace. The residual channel attention network serving as the super-resolution module is illustrated in dashed box. To guarantee the reconstruction quality, the module is added to the relatively larger scale feature layer $OP_2$.}
\label{fig22}
\end{figure}

The network architecture of EfficientSRFace is shown in Fig.~\ref{fig22}. To enhance the expression capability of degraded low-resolution image features and improve the detection accuracy of blurred faces, the feature-level image super-resolution reconstruction module illustrated in dashed box is incorporated into EfficientFace in which EfficientNet-B4 is used as the backbone. Considering that the image super-resolution reconstruction result largely depends on features with sufficient representation capability, the image super-resolution module is added to the feature layer $OP_{2}$ of the EfficientFace, since the scale of the feature map at $OP_{2}$ is 1/4 of the original scale after image pre-processing, and encodes abundant visual information to guarantee accurate super-resolution reconstruction.


\subsection{Image Super-Resolution Enhancement}\label{sec32}

Although EfficientFace~\cite{efficientface} achieves desirable detection accuracy when handling larger scale faces, it reports inferior performance when detecting low-resolution faces in the degraded image. In particular, numerous small faces carry much less visual clues, which makes the detector fail to discriminate them and increases the detection difficulty especially when the degraded images are not clearly captured. Consequently, we introduce Residual Channel Attention Network (RCAN)~\cite{rcan} within our EfficientSRFace, such that we perform feature-level super-resolution on EfficientFace for feature enhancement.

\begin{figure}
\centering
\includegraphics[scale=0.4]{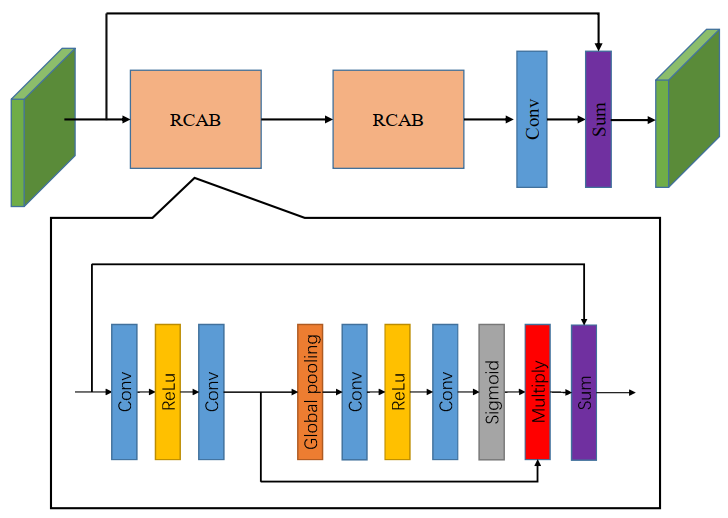}
\caption{Network structure of the RG module.}
\label{fig33}
\end{figure}

As shown in Fig.~\ref{fig22}, Residual Group (RG) component is used to increase the depth of the network and extract high-level features. It is also a residual structure which consists of two consecutive residual Channel Attention Blocks (RCABs). RCAB in Fig.~\ref{fig33} aims to combine the input features and the subsequent features prior to channel attention. Thus, it helps to increase the channel-aware weights and benefits the subsequent super-resolution reconstruction. Increasing the number of RGs contributes to further performance gains, whereas inevitably leads to excessive model parameters and computational overhead. This also increases the training difficulty of our network. For efficiency, the number of RGs is set to 2 in our scenario. Afterwards, the input low-level and high-level features resulting from RGs are fused by pixelwise addition strategy to enrich the feature information and help the network to boost the reconstruction quality. Finally, the upsampling strategy is used to increase the scale of features with the upsampling factor within our model set as 4. 

It should be noted that the RCAN module only plays a supplementary and auxiliary role during the training process, and it is discarded during reference without affecting detection efficiency within our EfficientSRFace. 

\subsection{Loss function}\label{sec33}

Mathematically, the overall loss function of our EfficientSRFace model is formulated as Eq.~(\ref{eq11}) which consists of three terms respectively calculating classification loss, regression loss and super-resolution reconstruction loss. Considering that our main focus is accurate detection, we assume the former two terms outweigh the SR loss and utilize the parameter $\varphi$ to balance the contribution of super-resolution reconstruction to the total loss function.

\begin{equation}
\label{eq11}
L_{ef}=L_{focal}+ L_{smooth}+\varphi L_{sr}
\end{equation}


More specifically, taking into account the sample imbalance, focal loss~\cite{focalloss} is utilized for the classification loss indicated as:

\begin{equation}
\label{eq22}
L_{focal}=-\alpha_{t}(1-p_{t})^{\gamma}log(p_{t})
\end{equation}
where $p_{t}\in \left [0,1  \right ]$ is the probability estimated for the class with label 1, and $\alpha_{t}$ is the balancing factor. Besides, $\gamma$ is the focusing parameter that adjusts the rate at which simple samples are downweighted.

In addition, smooth $\ell_1$ is used as the regression loss for accurate face localization as follows:

\begin{equation}
\label{eq33}
smooth_{\ell_1}(x)= \left \{
\begin{array}{ll}
    0.5x^{2} \quad\quad \lvert y \rvert <1 \\
    \lvert y \rvert -0.5 \quad otherwise
\end{array}
\right.
\end{equation}

In terms of super-resolution reconstruction loss, $\ell_1$ loss is adopted to measure the difference between the super-resolution reconstructed image and the target image formulated as follows:

\begin{equation}
    \label{eq44}
    L_{sr}=\frac{1}{W H}\sum_{i=1}^{W}\sum_{j=1}^{H}|y_{ij}-Y_{ij}|
\end{equation}
where $W$ and $H$ respectively represent the width and the height of the input image, while $y$ and $Y$ respectively represent the pixel values of the reconstructed and the target image.

\section{Experiments}\label{sec4}
In this section, extensive experiments are conducted to evaluate our proposed EfficientSRFace. Firstly, the public benchmarking datasets and experimental setting will be briefly introduced in our experiments. Next, comprehensive evaluations and comparative studies are also carried out with detailed model analysis.

\subsection{Datasets and evaluation metrics}\label{subsec41}
We have evaluated our EfficientSRFace network on four public benchmarking datasets for face detection including AFW~\cite{afw}, Pascal Face~\cite{pascalface}, FDDB~\cite{fddb} and WIDER Face~\cite{widerface}. Known as the most challenging large-scale face detection dataset thus far, WIDER Face comprises 32K+ images with 393K+ annotated faces exhibiting dramatic variances in scales, occlusion and poses. It is split into training (40\%), validation (10\%) and testing sets (50\%). Depending on different difficulty levels, the whole dataset is divided into three subsets, namely Easy, Medium and Hard subsets. For performance measure, Average Precision (AP) and Precision-Recall (PR) curves are used for metrics in different datasets.


\subsection{Implementation Details}\label{sec42}
In implementation, the anchor sizes used in our EfficientSRFace network are empirically set as \{16, 32, 64, 128, 256, 512\} and their aspect ratios are unanimously 1:1. In terms of the model optimization, AdamW algorithm is used as the optimizer and ReduceLROnPlateau attenuation strategy is employed to adjust the learning rate which is initially set to $10^{-4}$. If the loss function stops descending within three epochs, the learning rate will be decreased by 10 times and eventually decay to $10^{-8}$. The batch size is set as 4 for network training. The training and inference process are completed on a server equipped with a NVIDIA GTX3090 GPU under PyTorch framework.

\begin{table}[h] \addtolength{\textwidth}{0.1cm} \renewcommand{\arraystretch}{1.1}
\centering
\begin{center}
\caption{The accuracy and efficiency of different backbone network.}
\label{tab22}
\setlength{\tabcolsep}{1mm}{
\begin{tabular}{|c|c|c|c|c|c|c|}
\hline
Backbone & model& Easy  & Medium & Hard & Params(M) & MACs(G)\\
\hline
\multirow{2}{*}{EfficientNet-B0}&EfficientFace& 91.0\%& 89.1\%& 83.6\%&3.89&4.80\\     
&EfficientSRFace & 92.5\%& 90.7\%& 85.8\%&3.90&5.17\\
\hline
\multirow{2}{*}{EfficientNet-B1}&EfficientFace  & 91.9\%& 90.2\%& 85.1\%&6.54&7.81\\
&EfficientSRFace  & 92.7\%& 90.9\%& 86.3\%&6.56&8.43\\
\hline
\multirow{2}{*}{EfficientNet-B2}&EfficientFace  & 92.5\%& 91.0\%& 86.3\%&7.83&10.49\\
&EfficientSRFace  & 93.0\% & 91.7\% & 87.2\%&7.86&11.44\\
\hline
\multirow{2}{*}{EfficientNet-B3}&EfficientFace  & 93.1\%& 91.8\%& 87.1\%&11.22&18.28\\
&EfficientSRFace  & 93.7\%& 92.3\%& 87.6\%&11.27&20.06\\
\hline
\multirow{2}{*}{EfficientNet-B4}&EfficientFace  & 94.4\%& 93.4\%& 89.1\%&18.75&32.54\\
&EfficientSRFace  & \textbf{95.0\%}& \textbf{93.9\%}&\textbf{89.9\%}&18.84&35.83\\
\hline
\end{tabular}}
\end{center}
\end{table}

\subsection{\textbf{Data enhancement}}\label{sec43}
In terms of training the image super-resolution reconstruction module, the super-resolution labels are the original images, while the images preprocessed by random blur are delivered to the module. More specifically, in addition to the usual image enhancement methods such as contrast and brightness enhancement, random cropping and horizontal flip, we also leverage random Gaussian blur processing for the input images.


\subsection{Results}\label{sec44}
\subsubsection{Performance of using different backbones}

Table~\ref{tab22} presents the comparison of the EffcientFace and our proposed EfficientSRFace with different backbone networks. It can be observed EfficientSRFace consistently outperforms EfficientFace with different backbones used. In particular, when EfficientNet-B0 is used as the backbone, significant performance improvements of 1.5\%, 1.6\% and 2.2\% are reported on the three respective subsets. With the increase in the complexity of the backbone network structure, slightly declined performance gains can be observed. Since EfficientNet-B0 backbone has much less parameters and enjoys more efficient structure, it is prone to insufficient representation capability. In this sense, incorporating the feature-level super-resolution module is beneficial for enhancing the feature expression capability of the backbone, and bring more performance gains compared with our model using other efficient backbones. More importantly, the auxiliary super-resolution module incurs a small amount of additional parameters and slight growth in computational overhead, which suggests it hardly affects the detection efficiency in practical applications.

\begin{figure}
\centering
\includegraphics[scale=0.48]{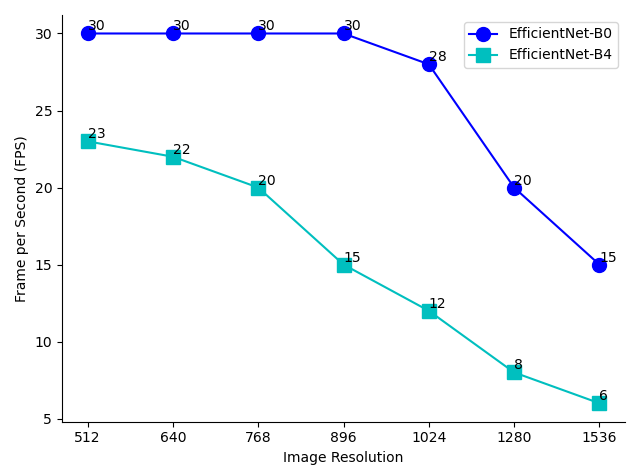}
\caption{FPS scores of our models at various scales using different backbones. For each image resolution using specific network backbone, the FPS score is obtained by averaging results of 1000 times.}
\label{fig55}
\end{figure}

In addition to the detection accuracy, we also present the Frame-Per-Second (FPS) values of our EfficientSRFace models for efficiency evaluation. As shown in Fig.~\ref{fig55}, although FPS generally exhibits a decreasing trend with the increase of image resolution, our model can still achieve real-time detection speed. For example, our model achieves 28 FPS speed for the image size of $1024 \times 1024$ when EfficientFace-B0 is used as backbone, which fully demonstrates the desirable efficiency of our EfficientSRFace.

\subsubsection{Parameter analysis of $\varphi$}

\begin{figure}
\centering
\includegraphics[scale=0.5]{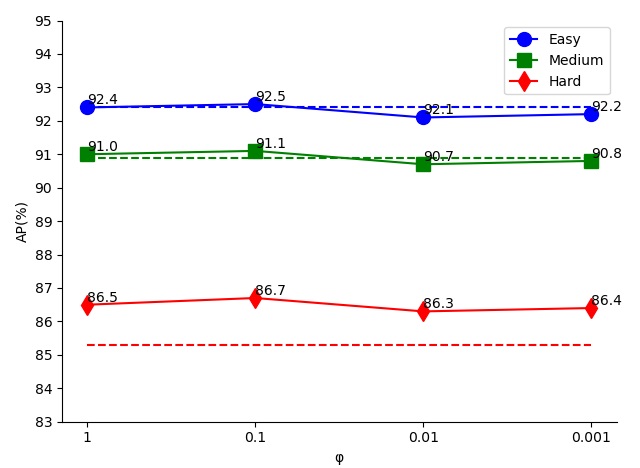}
\caption{Influence of different weight parameter values $\varphi$ on model detection performance, where the dotted line represents that no super resolution module is embedded.}
\label{fig44}
\end{figure}

As shown in Fig.~\ref{fig44}, we explore the effects of different weight parameter values of $\varphi$ on our model performance and compare the results with the EfficientFace (illustrated in dotted line). In this experiment, EfficientNet-B1 is used as the backbone network, and the batch size of model is set to 8. It can be observed that performance improvements to varying extents are reported on Hard subset with different $\varphi$ values. This demonstrates the substantial advantages of the super-resolution module particularly in the difficult cases including low-resolution face detection. Besides, the highest AP scores of 92.5\% (Easy), 91.1\% (Medium) and 86.7\% (Hard) are reported when $\varphi$ is set to 0.1, which is consistently superior to EfficientFace achieving 92.4\%, 90.9\% and 85.3\% on the three subsets. Thus, $\varphi$ is set to the optimal 0.1 in our experiments.

\subsubsection{Comparison of EfficientSRFace with state-of-the-art detectors}
In this part, the proposed EfficientSRFace is compared with state-of-the-art detectors in terms of both accuracy and efficiency on WIDER Face validation set. As shown in Table~\ref{tab33}, the competing models involved in our comparative studies include both heavy detectors such as DSFD and lightweight models like YOLOv5 variants and EXTD. In comparison to the heavy detectors, our EfficientSRFace-L using EfficientNet-B4 as the backbone achieves competitive performance with significantly reduced parameters and computational costs. In particular, EfficientSRFace-L reports respective 95.0\%, 93.9\% and 89.9\% AP scores on Easy, Medium, and Hard subsets, which is on par with DSFD achieving 96.6\%, 95.7\% and 90.4\% accuracies. However, our model enjoys approximately 6× reduced parameters and costs 10× decreased MACs. Particularly, when EfficientNet-B0 is used as the backbone in our EfficientSRFace detector, EfficientSRFace-S achieves preferable efficiency which is competitive with lightweight YOLOv5n, while outperforming the latter by 5\% on Hard set. Benefiting from the efficient architecture design of EfficientFace, our model enjoys different variants ranging from extremely efficient model superior to the other lightweight competitors and the relatively larger network comparable to some heavyweight models.

\begin{table}[h] \addtolength{\textwidth}{-0.1cm} \renewcommand{\arraystretch}{1.1}
\centering
\caption{Comparison of EfficientSRFace and other advanced face detectors. EfficientSRFace-S and EfficientSRFace-L denote our two models with EfficientNet-B0 and EfficientNet-B4 respectively used as the backbones.}
\label{tab33}
\setlength{\tabcolsep}{1mm}{
\begin{tabular}{|c|c|c|c|c|c|}
\hline
Model & Easy  & Medium & Hard & Params(M) & MACs(G)\\
\hline
MogFace-E~\cite{mogface}&\bfseries97.7\% &\bfseries96.9\% & 92.01\% & 85.67	& 349.14\\
MogFace~\cite{mogface}&97.0\%&96.3\%	&\bfseries93.0\%& 85.26	& 807.92\\
AlnnoFace~\cite{alnnoface}	&97.0\%	&96.1\%	& 91.8\%  & 88.01	& 312.45\\
SRNFace-1400~\cite{srnface}	&96.5\%	&95.2\%	& 89.6\%  & 53.38 & 251.94\\
SRNFace-2100~\cite{srnface}&96.5\%&95.3\%&90.2\%& 53.38& 251.94\\
DSFD~\cite{dsfd}	&96.6\%	&95.7\%	& 90.4\%  & 120	  & 345.16\\
yolov5n-0.5~\cite{yolo5face}&90.76\%&88.12\%& 73.82\% & 0.45  & 0.73\\
yolov5n~\cite{yolo5face}	 &93.61\%&91.52\%& 80.53\% & 1.72  & 2.75\\
yolov5s~\cite{yolo5face}	 &94.33\%&92.61\%& 83.15\% & 7.06  & 7.62\\
yolov5m~\cite{yolo5face}	 &95.30\%&93.76\%& 85.28\% & 21.04 & 24.09\\
yolov5l~\cite{yolo5face}&95.78\%&94.30\%& 86.13\% & 46.60 & 55.31\\
EXTD-32~\cite{extd}&89.6\%&88.5\%	& 82.5\%  & 0.063 & 5.29\\
EXTD-64~\cite{extd}&92.1\%&91.1\%	& 85.6\%  & 0.16  & 13.26\\
EfficientSRFace-S~(Ours)& 92.5\%&90.7\%&85.8\%&3.90&5.17\\
EfficientSRFace-L~(Ours) &95.0\%	&93.9\%	& 89.9\%  & 18.84 & 35.83\\
\hline
\end{tabular}}
\end{table}

\subsubsection{Comprehensive Evaluations on the four Benchmarks}
\begin{figure}
\subfigure[AFW]{
\includegraphics[width=1.4in,height=1.2in]{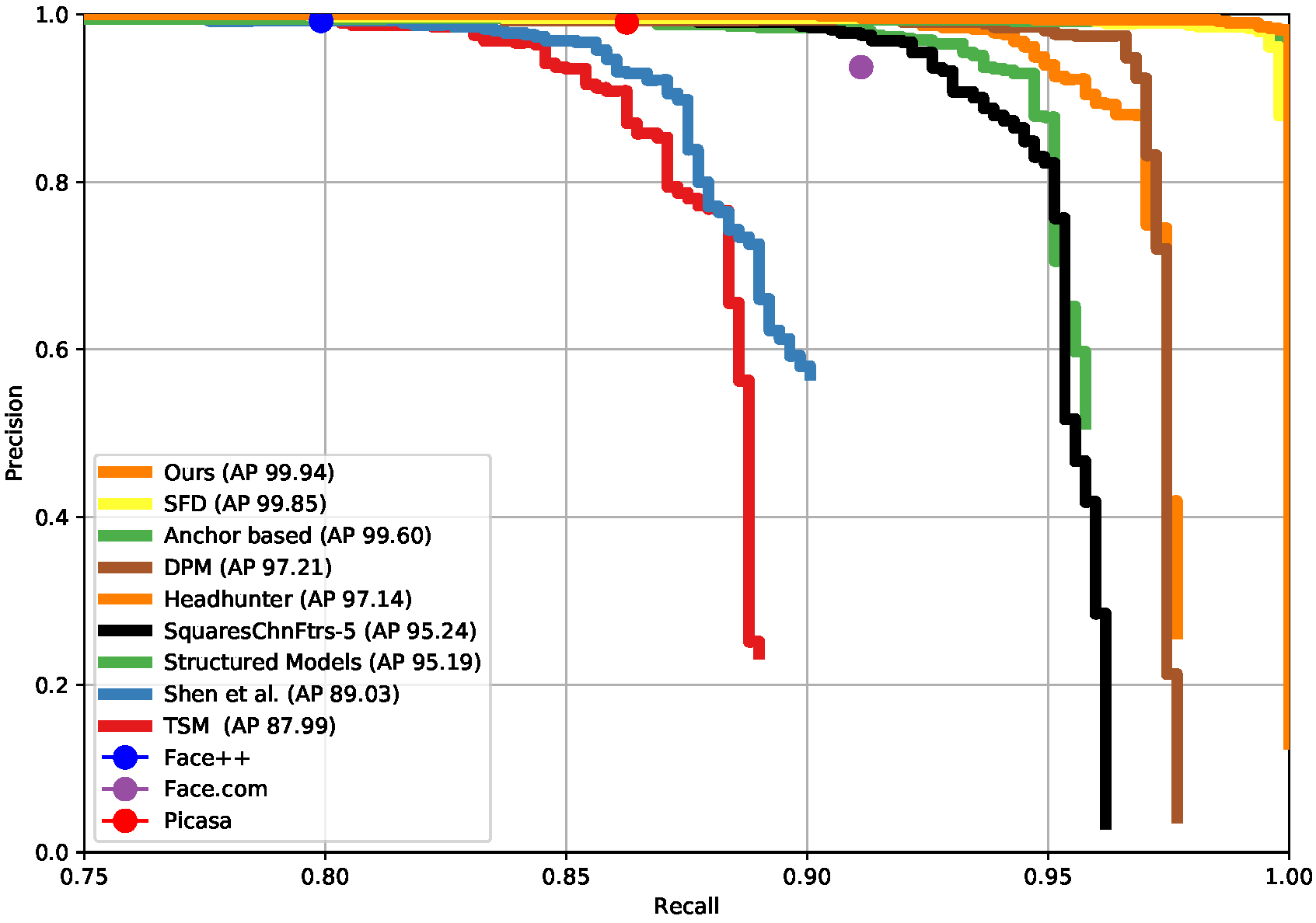}
}
\subfigure[Pascal Face]{
\includegraphics[width=1.4in,height=1.2in]{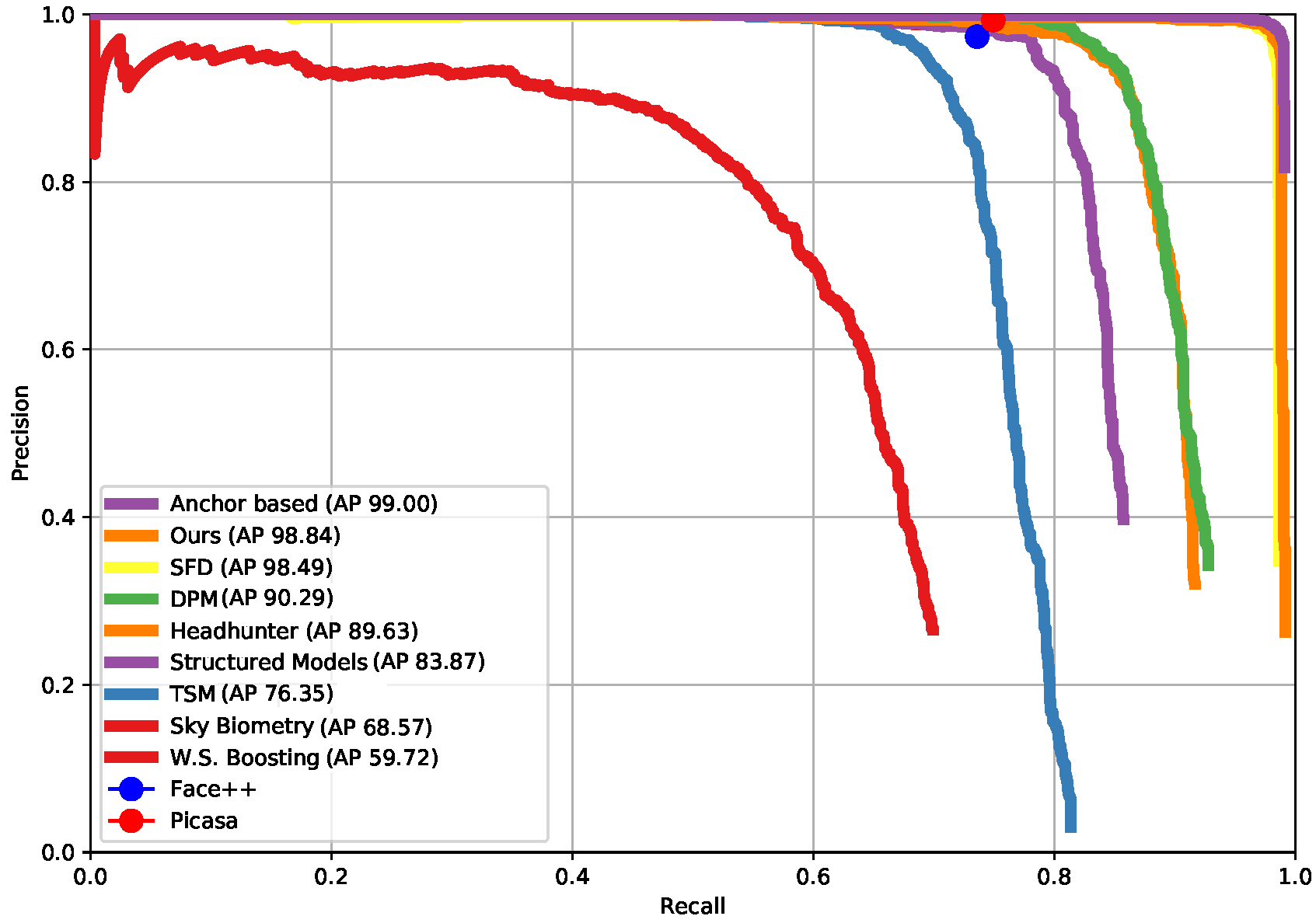}
}
\subfigure[FDDB]{
\includegraphics[width=1.4in,height=1.2in]{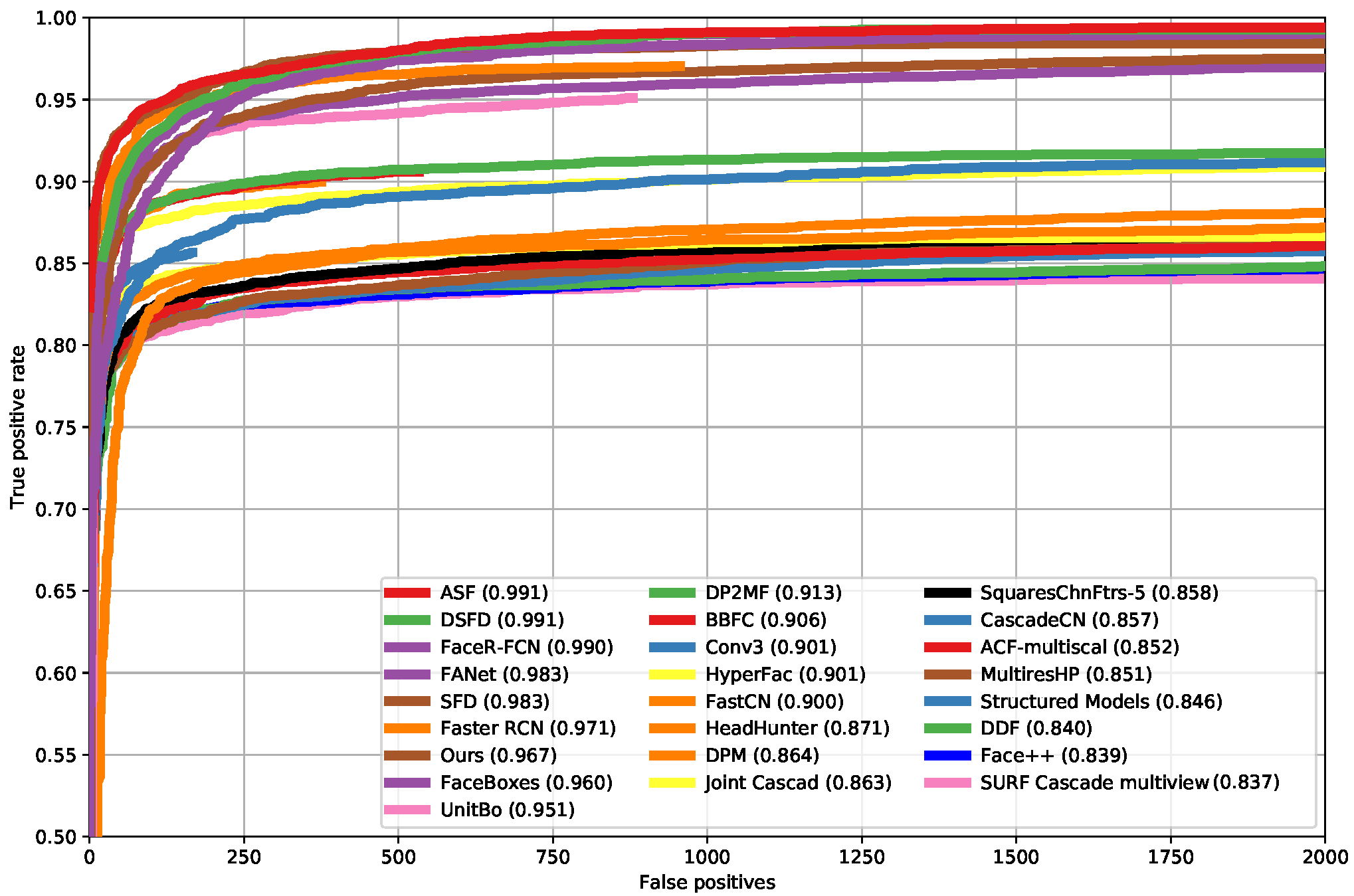}
}
\caption{Evaluation on common face detection datasets.}
\label{fig66}
\end{figure}

\begin{figure}
\subfigure[Easy]{
\includegraphics[width=1.4in,height=1.1in]{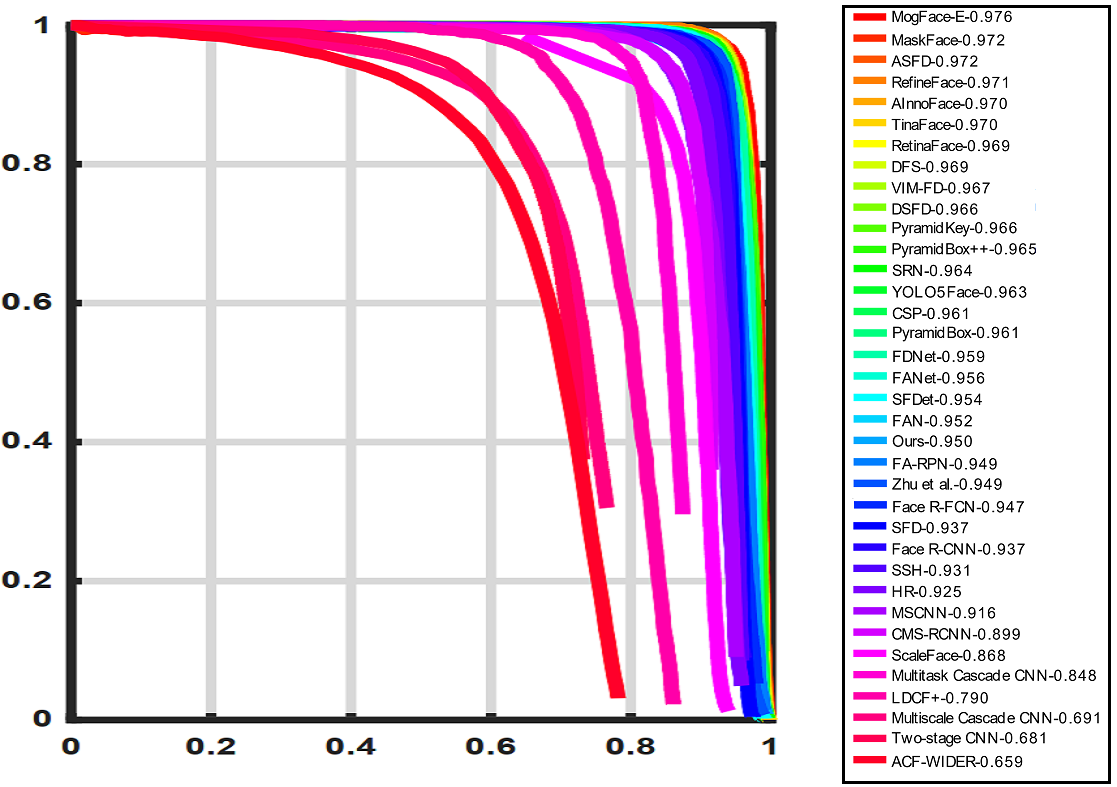}
}
\subfigure[Medium]{
\includegraphics[width=1.4in,height=1.1in]{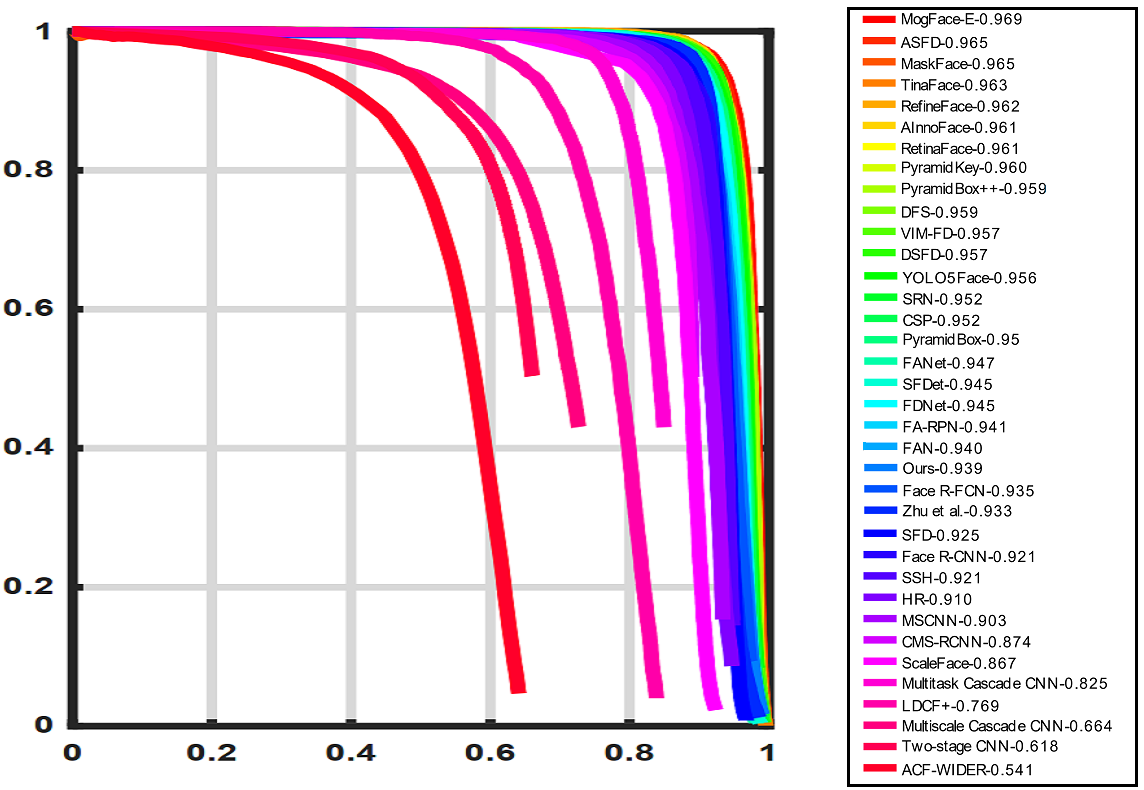}
}
\subfigure[Hard]{
\includegraphics[width=1.4in,height=1.1in]{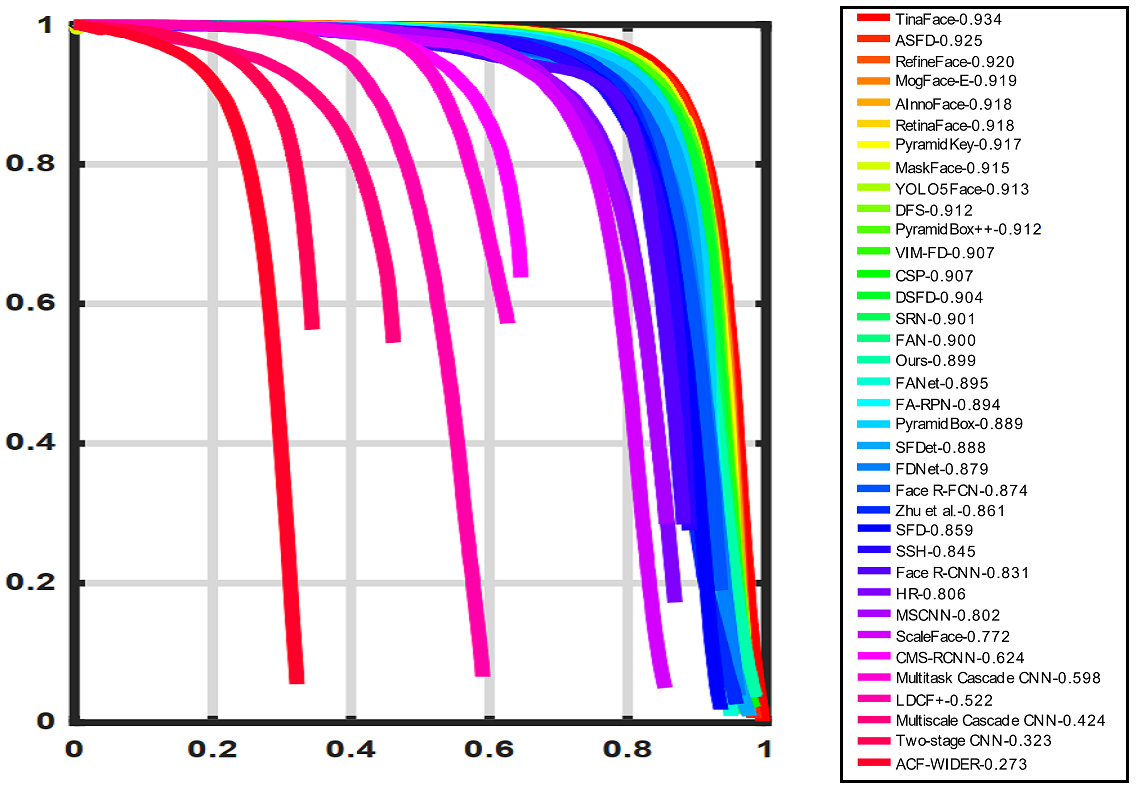}
}
\caption{PR curves of different methods on validation set of WIDER Face dataset.}
\label{fig77}
\end{figure}

In this section, we will comprehensively compare EfficientSRFace and other advanced detectors in the four public datasets. Fig.~\ref{fig77} shows precision-recall (PR) curves obtained by different models on validation set of WIDER Face dataset. Although EfficientSRFace is still inferior to some advanced heavy detectors, it still achieves competitive performance with promising model efficiency. In addition to WIDER Face dataset, we also evaluate our EfficientSRFace on the other three datasets and carry out more comparative studies. As shown in Table~\ref{tab44}, our EfficientSRFace-L achieves respective 99.94\% and 98.84\% AP scores on AFW and PASCAL Face datasets. In particular, EfficientSRFace consistently beats the other competitors including even heavyweight models like MogFace and RefineFace~\cite{refineface} on AFW. In addition to AP scores, we also present PR curves of different detectors on AFW, PASCAL Face and FDDB datasets as shown in Fig.~\ref{fig66}. On FDDB dataset, more specifically, when the number of false positives is 1000, our model reports the true positive rate up to 96.7\%, surpassing most face detectors.

\begin{table}[h] \addtolength{\textwidth}{0.0cm} \renewcommand{\arraystretch}{1.1}
\begin{center}
\caption{Comparison of EfficientSRFace and other detectors on the AFW and PASCAL Face dataset (AP).}
\label{tab44}
\setlength{\tabcolsep}{6mm}{
\begin{tabular}{|c|c|c|}
\hline
	   Models & AFW	& PASCA Face\\
\hline
RefineFace~\cite{refineface} & 99.90\%	& \bfseries99.45\%\\
FA-RPN~\cite{farpn} 	  & 99.53\%	& 99.42\%\\
MogFace~\cite{mogface}	  & 99.85\%	& 99.32\%\\
SFDet~\cite{sfdet}       & 99.85\%	& 98.20\%\\
SRN~\cite{srnface}	      & 99.87\%	& 99.09\%\\
FaceBoxes~\cite{faceboxes}  & 98.91\%	& 96.30\%\\
HyperFace-ResNet~\cite{hyperface}	& 99.40\%	& 96.20\%\\
STN~\cite{stn}	      & 98.35\%	& 94.10\%\\
Ours	              & \bfseries99.94\%& 98.84\%\\
\hline
\end{tabular}}
\end{center}
\end{table}

\section{Conclusions}\label{sec5}
In this paper, we develop an efficient network architecture based on EfficientFace termed EfficientSRFace to better handle the low-resolution face detection. To this end, we embed a feature-level super-resolution reconstruction module to feature pyramid network for enhancing the feature representation capability of the model. This module plays an auxiliary role in the training process and can be removed during the inference without increasing the inference time. More importantly, this supplementary role incurs a small amount of additional parameters and limited growth in computational overhead without damaging model efficiency. Extensive experiments on public benchmarking datasets demonstrate that the embedded image super-resolution module can significantly improve the detection accuracy at a small cost.

\end{document}